\documentclass[12pt]{spieman}  
\usepackage{amsmath,amsfonts,amssymb}
\usepackage{graphicx}
\usepackage{setspace}
\usepackage{tocloft}
\usepackage[utf8]{inputenc}
\usepackage{svg}
\usepackage{listings}
\usepackage{enumitem}
\usepackage{color}
\usepackage{nth}
\usepackage[style=base]{caption}

\title{\textit{leave a trace} -- A People Tracking System Meets Anomaly Detection}

\author[a]{Dominik Rueß}
\author[a]{Konstantinos Amplianitis}
\author[a]{Niklas Deckers}
\author[a]{Michele Adduci}
\author[b]{Kristian Manthey}
\author[a]{Ralf Reulke}
\affil[a]{Humboldt-Universität zu Berlin, Institut für Informatik, Unter den Linden 6, 10099 Berlin, Germany}
\affil[b]{German Aerospace Center, Rutherfordstraße 2, 12489 Berlin, Germany}

\cftpagenumbersoff{figure}
\cftpagenumbersoff{table} 

\begin{document} 
\maketitle

\begin{abstract}
{Video surveillance always had a negative connotation, among others because of the loss of privacy and because it may not automatically increase public safety. If it was able to detect atypical (i.e. dangerous) situations in real time, autonomously  and anonymously, this could change. A prerequisite for this is a reliable automatic detection of possibly dangerous situations from video data. This is done classically by object extraction and tracking. From the derived trajectories, we then want to determine dangerous situations by detecting atypical trajectories. However, due to ethical considerations it is better to develop such a system on data without people being threatened or even harmed, plus with having them know that there is such a tracking system installed. Another important point is that these situations do not occur very often in real, public CCTV areas and may be captured properly even less. \\
In the artistic project \textit{leave a trace} the tracked objects, people in an atrium of a institutional building, become actor and thus part of the installation. Visualisation in real-time allows interaction by these actors, which in turn creates many atypical interaction situations on which we can develop our situation detection. The data set has evolved over three years and hence, is huge. \\
In this article we describe the tracking system and several approaches for the detection of atypical trajectories.}
\end{abstract}


{\noindent \footnotesize\textbf{Contact Address:}  \linkable{reulke@informatik.hu-berlin.de} }

\begin{spacing}{2}   

\section{Introduction}
\label{sect:intro}

\textit{Leave a trace} is an object detection and tracking system that was built on existing state-of-the-art algorithms and improved for the purpose of tracking people walking in an atrium of the \textit{Charité -- Universitätsmedizin Berlin}, in Berlin, Germany. It was launched in early 2014 and has been running every since. The tracks (trajectories) of one or more people are displayed in real-time on a large monitor placed in the atrium of the main building. The visitors can interact with their own track to create distinct imagery. The tracks will accumulate over time but also slowly fade away to avoid cluttering of the screen and adding the time component. The system resets itself every morning as everyday it has its own collection of tracks. A typical output of the screen may consist of cumulative tracks for the usual walking paths. Every now and then a visitor may write or draw something, which is -- in a sense -- some kind of unusual behaviour.

Apart from the artistic nature of this project, it also provides a large database of anonymized movement, collected every day -- over the years. This data can be used to develop methods to analyse the movement and to detect anomalous behaviour, for instance.
This is especially true since it has users react to their own movement being displayed and hence, acting differently to most other visitors who usually just walk to their rooms and back.

The system works with a CCTV camera placed almost vertical in the \nth{5} floor in the atrium, to reduce perspective effects.
The high data throughput of 15 frames per second, each frame of size 2 mega-pixels, is transmitted via Gig-E to a server which runs the image processing, tracking, data visualisation and data storage.
The screen has a diagonal of 80 inches and a resolution of 4K and thus also required a high throughput graphics card and connection.

In this article we will describe the technical and software setup of the system. The contribution is organized as follows: 
Firstly, we describe the artistic goal of this project.
Secondly, we will provide insight into the hardware required to work on the huge amount of input data, as well as to work interruption free.
Thirdly, we will list the methods used to extract and track the movement of people from the input images.
Lastly, we present an evaluation method to detect anomalies in trajectories. 

\begin{figure}[t!] 
\begin{center}  
\includegraphics[width=0.8\textwidth]{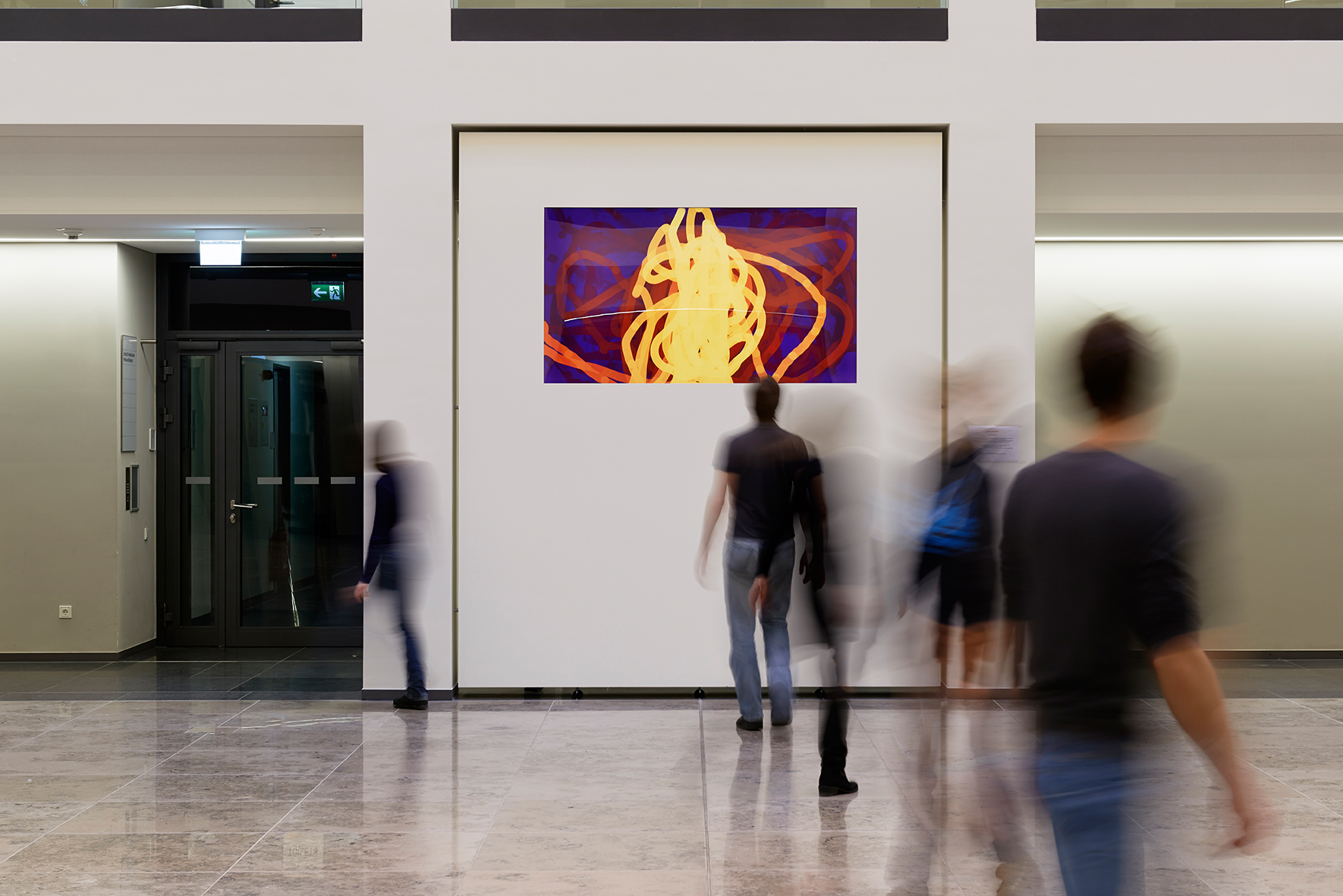}
\end{center}
\caption {
\label{fig:screenexample}
View on the installation of the \textit{leave a trace} system.}
\end{figure}

The Internet address of the project is \linkable{https://leave-a-trace.charite.de/leave\_a\_trace/}. It provides an on-line and live visualisation of the current movement in the atrium. Furthermore, it provides recent daily data sets which you can use for projects of your own. An example python script can be used for basic data reading (download here: \linkable{https://sourceforge.net/projects/leave-a-trace-tools/}). 
In \autoref{fig:screenexample} the installation and the screen is shown.

\section{Related Work}

\subsection{Sensors and Systems}

Pedestrian tracking from video stream in urban environment is often used for safety and security reasons. Multi-camera multi-target tracking is the typical approach, which is also one of the most active research topics in computer vision. 

In recent years, robust, real-time and model-based tracking techniques have been developed for rigid and non-rigid moving objects. They enable automated monitoring and event detection. 

In general, there is the following approach: object detection, tracking and fusion, as well as the interpretation of the derived trajectories into normal or typical and abnormal or atypical. In principle, the differentiation of abnormal trajectories is also of interest.

Pedestrian detection can be carried out either by stationary or by moving cameras (on-board, e.g. Ikoma et al\cite{RN20} and off-board camera camera, e.g. Aycard et al\cite{RN37}). To avoid occlusions and extend the field of view of the whole camera system, multi-camera systems are used (e.g. see Batista\cite{RN41}, Berclaz et al\cite{RN3} or Fuhr and Jung\cite{RN19}). In addition to classical cameras, which operate in the visible spectral range pan-chromatically or colorfully in rgb, NIR (e.g. Kumar\cite{RN23}) and thermal infrared cameras (e.g. Li et al\cite{RN9}) are also used.

Camera calibration is also an important point in the use of multi-camera systems and cameras with a special imaging geometry (see Ikoma\cite{RN26} where a omni-directional camera was used). The aim is to avoid the complex camera calibration, especially for multi-camera systems (e.g. Varga et al\cite{RN17} and Hyodo et al\cite{RN35}).

The use of alternative sensors and the fusion with cameras or stereo cameras is also possible. Streubel and Yang\cite{RN14} reports about fusion of stereo camera and MIMO-FMCW radar, Ikoma\cite{RN26} uses a composite sensor of laser range finder and omni-directional camera.

\subsection{Detection of Abnormal Behaviour}

Berclaz et al\cite{RN3} introduce a new behavioural model to describe pedestrian movements. These are capable of discerning movement patterns resulting from the mixture of different categories of random trajectories. It is also capable of characterizing atypical individual movements.

Gong\cite{RN4} presents  the  results of a trainable stochastic temporal models for automatic event and behaviour detection. They provide an approach for modelling and recognizing complex activities with simultaneous movement of several objects. Dynamic probabilistic graph models are used to model the temporal relationships between a set of different object-time events. 

The paper of Junejo\cite{RN1} deals with the problem of scene modelling. A scene model is constructed to distinguish normal and acceptable behaviour from atypical one. The proposed method is divided into a training phase and a test phase. During the training phase, the input trajectories are used to identify different paths or routes. Important discriminating features are then extracted to learn a dynamic Bayesian network (DBN).

The paper by Makris and Ellis\cite{RN5} deals with the problem of extracting frequently used pedestrian paths. Path models are trained from the accumulation of trajectory data over long periods of time. In particular, the model can be used to calculate a probabilistic prediction of the position of the pedestrian ahead of time and to support the recognition of unusual behaviour.

Reulke et al\cite{RN2} discussed a trajectory-based recognition algorithm to extend the common approaches for atypical event detection in multi-object traffic scenarios. They have searched for path-based types of information (e.g. maps of velocity patterns, trajectory curvatures, or unpredictable movements). They also implemented the merging of object data from different cameras with a multi-target tracking approach. This approach opens up possibilities to identify and specify traffic objects, their location, speed and other characteristic object information.

A statistically substantiated model of object trajectories is presented by Johnson and Hogg\cite{RN44}. Trajectory data is provided by a tracker based on Active Shape Models, from which a model of the distribution of typical trajectories is trained. An approach for identifying atypical events based on the model described is given.

The article by Zhou et al\cite{RN45} investigates the problem of detecting abnormal events by analysing the motion paths in videos. The core of the problem is to provide a robust and accurate function for measuring the similarities of trajectory pairs.

Most of the articles only consider individual trajectories. In Reulke\cite{reulke1} an approach to finding interactions between objects in a scene based on trajectories has been introduced.

\section{Goal of the Project}
The project is part of an art in architecture installation inside an atrium of a building of the Charité in Berlin, Germany: the CharitéCrossOver building (CCO). Tyyne Claudia Pollmann developed a concept of a movement visualisation for this atrium. Persons' walking paths would be represented in real-time on a large screen, in different colors for each day: The track color of a day becomes the background color of the next day. For each week, a new random color collection is selected. The line width represents the day of the week, it increases from thin lines on Mondays to very thick lines on Sundays. Tracks will disappear with time to be able to recognise recent activities.  People moving in the atrium are the originators of the traces and thus the essential part of the art. Every day, apart from the usual tracks being walked, individual traces are being produced, which are recognisable as certain shapes, symbols or messages. The atrium is frequently used for science exhibitions, conferences or other types of gatherings, which result in different and specific patterns due to the spatial installations for the events. So, the outcome is a variety of different patterns and traces which form a constantly changing visual information on the screen.

\section{Hardware Description}
The CCTV-Camera is a fast and robust industrial camera with a sensor size of 2 mega-pixel and a high quality lens. We have calibrated this Camera using the Brown model\cite{brown}, to correct for distortion and principal point offset. This model is used frequently in photogrammetric applications. It does correct distorted image coordinates for barrel, pincushion and trapezoidal errors. This allows for a much more accurate map of the image plane to the ground plane and hence, a more accurate tracking of objects. A high position of the camera reduces the effects of perspective projection, which will influence object positions on the ground plane mapping. The camera is connected to the computation server by an Gigabit-Ethernet connection, such that we could employ the GigE-Vision protocol for the large data throughput. 

For the computation unit we chose server components, such that the system will run with as few interruptions and failures as possible.  The most important system parts are an Intel Xeon E5 3.4GHz CPU with 8 virtual Cores, 16GB RAM and a NVIDIA GeForce GTX 670 graphics card, the latter for both, displaying in 4K and running image processing algorithms in parallel. An uninterrupted power supply will secure the system from monthly routine power cutoffs. The operating system is Debian and it allows for automation of daily routines, including re-start, emptying system cache, data backup, video generation, etc.

\autoref{fig:system} shows the different components in a block diagram. 

The above system has already exceeded the budget and hence it was not possible to install more cameras to improve distinguishing persons from the background -- e.g. by stereo methods.

\section{Image Processing and Tracking}
The people tracking system works similar to the methods described in Reulke et al and Rueß et al\cite{RN2,reulke1, ruess}, however it was altered and improved to be able to track in an indoor environment prone to changing lighting conditions, a high frame rate and large image size from only one camera. Also it was optimized to tracking people.

The image processing of up to 15 frames per second, of which each frame has 2 mega-pixels, is a very challenging task.
We have implemented a chain of tools which ultimately result in the tracks visualising the people's movement.
The goal was to be as exact and as fast as possible, this will always result in compromises regarding the resolutions and image parts we used.

Wherever possible we preferred the GPU methods of the computer vision library \textit{OpenCV} over the CPU ones. At the same time, whenever we encountered data parallelism in CPU methods, we have all 8 cores work on that particular task.

The steps of the proposed framework are given below:

\begin{figure}[t!] 
\begin{center}
\includegraphics[width=0.8\textwidth]{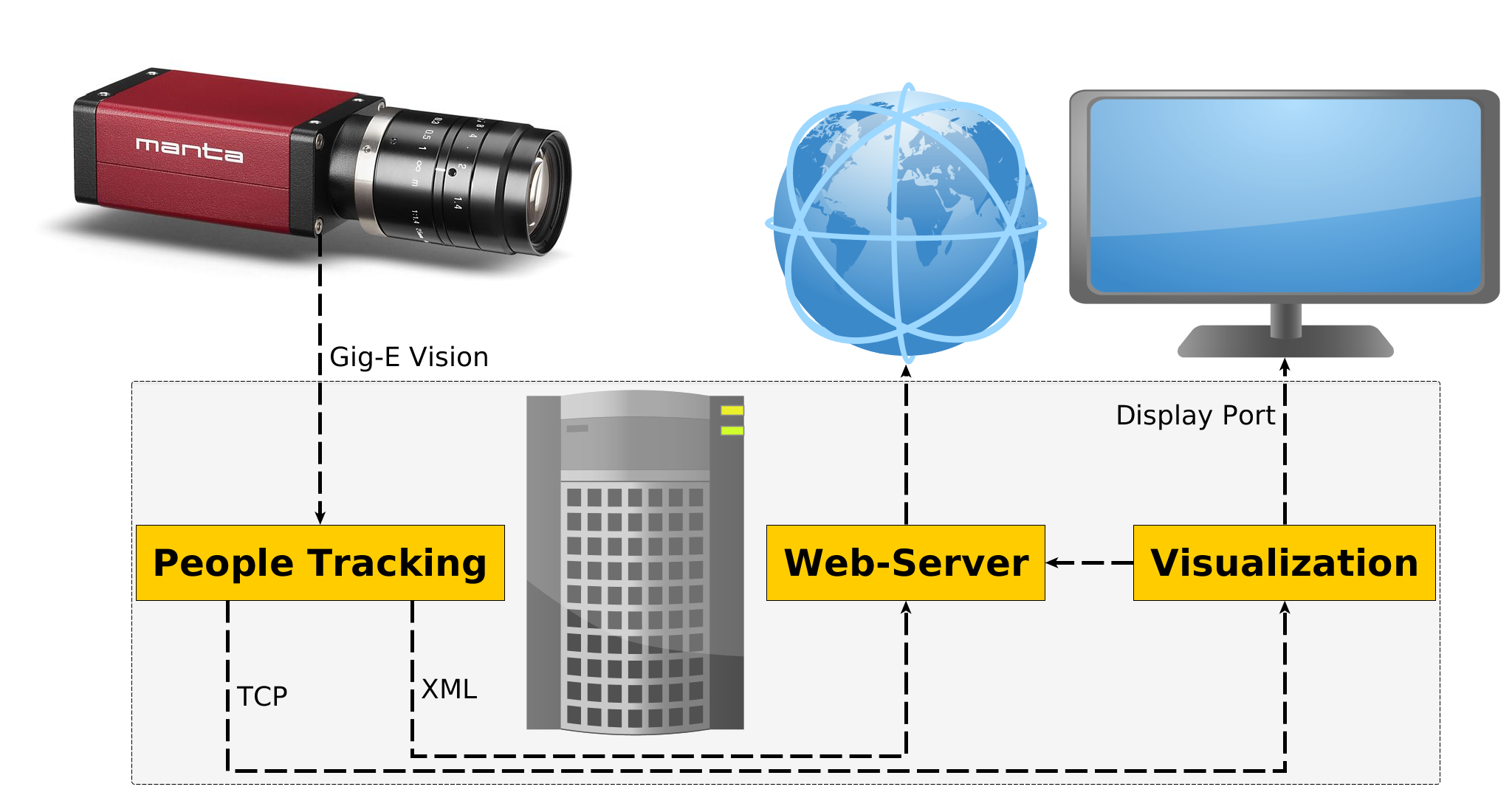}  
\end{center}
\caption 
{\label{fig:system}
Block diagram of the system with the interaction of the different components.}
\end{figure} 

\subsection{Pre-Processing}
First of all, it was important to introduce some plausibility checks. For instance, the image may be too dark or too bright due to fast changes in the illumination conditions. In this case we have the tracking stop for some frames. As mentioned above, we needed to detect sunlight areas. Connected areas with enough saturated pixels will be added to a blacklist mask. These areas will not be considered in the further object extraction. 

\subsection{Object Extraction}
Object detection was performed using a adapted background estimation approach. The reason for this is because the background would sometimes undergo significant changes e.g., an event in the atrium. Sun reflection or cloud shadows were also taken into account for modelling the background. 

The well-known open source \textit{OpenCV} library provides a Gaussian Mixture-based Model (GMM)\cite{pakorn} implementation, which is also available for GPUs. The advantage of this approach is that it can remove instant shadows due to the manually-defined number of frames that it requires to build the background. For example, for a person appearing in the scene for 5 seconds and an update of the background every 10 seconds, the person's shadow will not become part of the background.

Next step is the extraction of active blobs satisfying some criteria (not too thin, too large or too small). The barycentre of the blobs serve as object coordinates, which are later on undistorted and projected to ground floor coordinates.

All of this is done in planar space, by projecting the image coordinates onto the ground plane.

\subsection{Object update}
Before and after solving the assignment problem, it is necessary to reset the detection and tracking system. For instance, check old, likely dead tracks, even if they are within the usually active region. Also, if a new track would be initiated but there is already a recent nearby track, we try to connect these two. Sometimes, if a group of people splits within the active region, new tracks will be initiated. We try to add a track history for this new track from a recent and nearby older track. Again, some plausibility checks are necessary: For instance, some tracks have too large jumps.

For assigning blobs to existing tracks we use a Kalman filter to predict tracks, however the prediction is not always accurate, especially in the initialisation stage but also when the moving person turns abruptly, for instance. We added the \textit{OpenCV} optical flow method\cite{farneback} for an independent prediction of the object. This is implemented for the GPU but is not fast enough for the 2MP image. Therefore a scaled down version of this algorithm was developed.

The solution of the assignment problem is purely based on the two predicted positions. Due to the filtering of the blobs there are not too many objects left, so we can use the Hungarian algorithm\cite{hungarian1,hungarian2} to find the globally most plausible matching. 

For robustness' sake not every detection is used in the matching. New (possibly unmatched) detections will have to accumulate over a couple of frames to become a proper track. A filter also removes non-plausible detections, like not  within a mask, masked out by saturated (or dark) pixels, out of boundary and too extreme object size. 

\subsection{Data Preparation}
The data preparation turns out to be an expensive task too. Consider drawing a track which consists of many line segments:
For every corner a joint needs to be computed. Then, the drawing is not opaque, see also \autoref{fig:example}. All of this needs to be done in 4K/UHD resolution and with many and possibly long tracks, being repainted on every frame with different alpha. 

There are more steps for data preparation, once a second the web-server will obtain a new rendered and downscaled image.
A higher resolution image is saved at larger intervals, also provided for downloading. Finally, the data XML file will be updated quite frequently to not lose too much data in case of a system failure.

\section{Tracking System Evaluation}
\label{sec:trackingsystemevaluation}
\begin{figure}[t!]
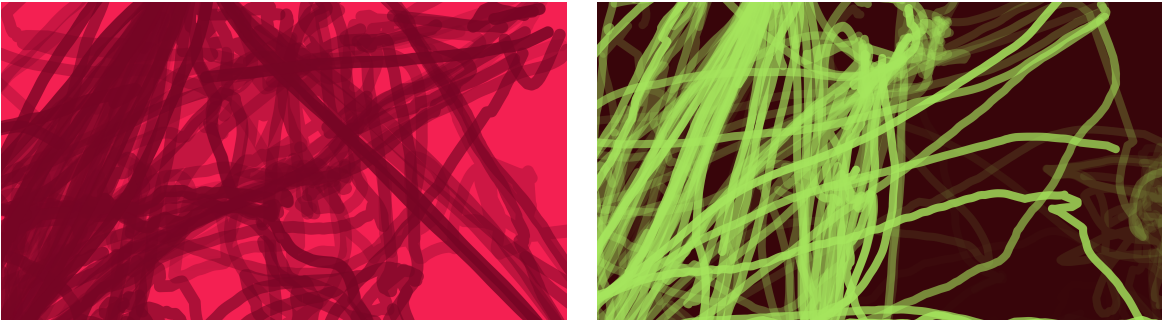
 
\begin{center}
\begin{tabular}{cc}
\includegraphics[width=7.5cm]{2015-02-06_tracks}
&\includegraphics[width=7.5cm]{2016-09-01_tracks.pdf}
\end{tabular}
\end{center}
\caption 
{ \label{fig:example}
{Exemplary (artistic) output of two different days. 
You can see an accumulation of tracks for typical walking paths. Note how the colors and line widths are different on a per-day basis. Refer to the main text for more information.}}
\end{figure}

\autoref{fig:example} provides some exemplary outputs for different days.
The different foreground, background colors and line widths are part of the artistic concept.

The tracking system works with a throughput of 12 frames per second, on average, each frame having a size of 2 mega-pixels.
It is able to track an arbitrary number of people which has been proven at days with conferences taking place in that atrium. It will track almost all of the people and other objects like cleaning machines etc.

We have encountered the following issues: Firstly, due to perspective and the swinging motion behaviour of humans, the tracks have a slight wobble in them, roughly to the extent of up to a third of a person size (viewed from above). See  \autoref{sec:dataevaluation} for how this issue is addressed in the evaluation of this data.

Secondly, the atrium is prone to sunlight changes, since they rays reflect from the walls and ultimately cast areas of bright light onto the floor. This light is so much brighter than the surroundings that we could only temporarily block these areas from tracking, wherever a collection of saturated pixels were found. The camera sensor does in fact have a dynamic range of 12 significant bits but that was just not enough.

Lastly, since the atrium is in some building within the Charité, people tend to wear bright, white coats more often. Thus, there are some issues with loosing track of people more often. This is due to the very bright floor tiles which are not very distinct from the cloak's white color. Again, the dynamic range is still not high enough to distinguish between these types of cloak and the floor.

\section{Data Evaluation -- Anomaly Detection}
\label{sec:dataevaluation}

\subsection{Justification}

In the above sections we have described how we obtained a top-down view of the scene with 2D trajectories of each person. An automatic image analysis with the aim of recognising atypical trajectories may be applied, within \textit{leave a trace} these are desired as a creative factor and determine the meaning of the installation. Image analysis frequently serves the recognition of atypical events in public space with the aid of optical sensors and are frequently correlated with hazardous situations. Atypical events can be extrapolated from untypical motion sequences. To this aim modern camera technology is used to (decentralized) capture situations, which are then centrally reconstructed into an overall situation and finally analyzed using intelligent-learning procedures.
Masking by infrastructure and other participants is problematic here. The selection (segmentation) of individual people or particular objects is complicated and sometimes even impossible. It has become apparent that neither algorithms nor monitoring equipment (e.g. stereo cameras) have reached adequate levels to be able to solve such problems reliably.
Nevertheless it must be said that bases for this were established within this project and realized prototypically. In this way movements can be detected, segmented from the background and monitored.

\autoref{fig:trajectories3} shows two more examples. The corresponding values are provided in \autoref{tab:Results}.
 
We pursued two approaches in order to attempt an evaluation. \autoref{fig:trajectories1} shows an example of trajectories recorded within a single day (left) and an extracted trajectory as an example.

\begin{table}[h!]
\centering
\begin{tabular}{lllllll} \hline
\textbf{ID}  & \textbf{nPoints} & \textbf{dFit}  & \textbf{dist}  & \textbf{cRect} & \textbf{cMain}  & \textbf{Figure} \\ \hline
4   & 178  & 11.2  & 2276  & 4589  & 52.9 & \autoref{fig:trajectories1} right \\
53  & 2590 & 342.7 & 26933 & 59978 & 1.95 & \autoref{fig:trajectories2} right\\
488 & 1129 & 160.1 & 5788  & 3305  & 0.99 & \autoref{fig:trajectories3} left\\
541 & 923  & 292.1 & 7558  & 7093  & 1.34 & \autoref{fig:trajectories3} right\\
\hline
\end{tabular}
\caption{Parameters for normal (ID 4) and atypical trajectories.}
\label{tab:Results}
\end{table}

\subsection{First Approach}

\begin{figure}[htbp] 
\centering
\fbox{\includegraphics[width=0.45\textwidth]{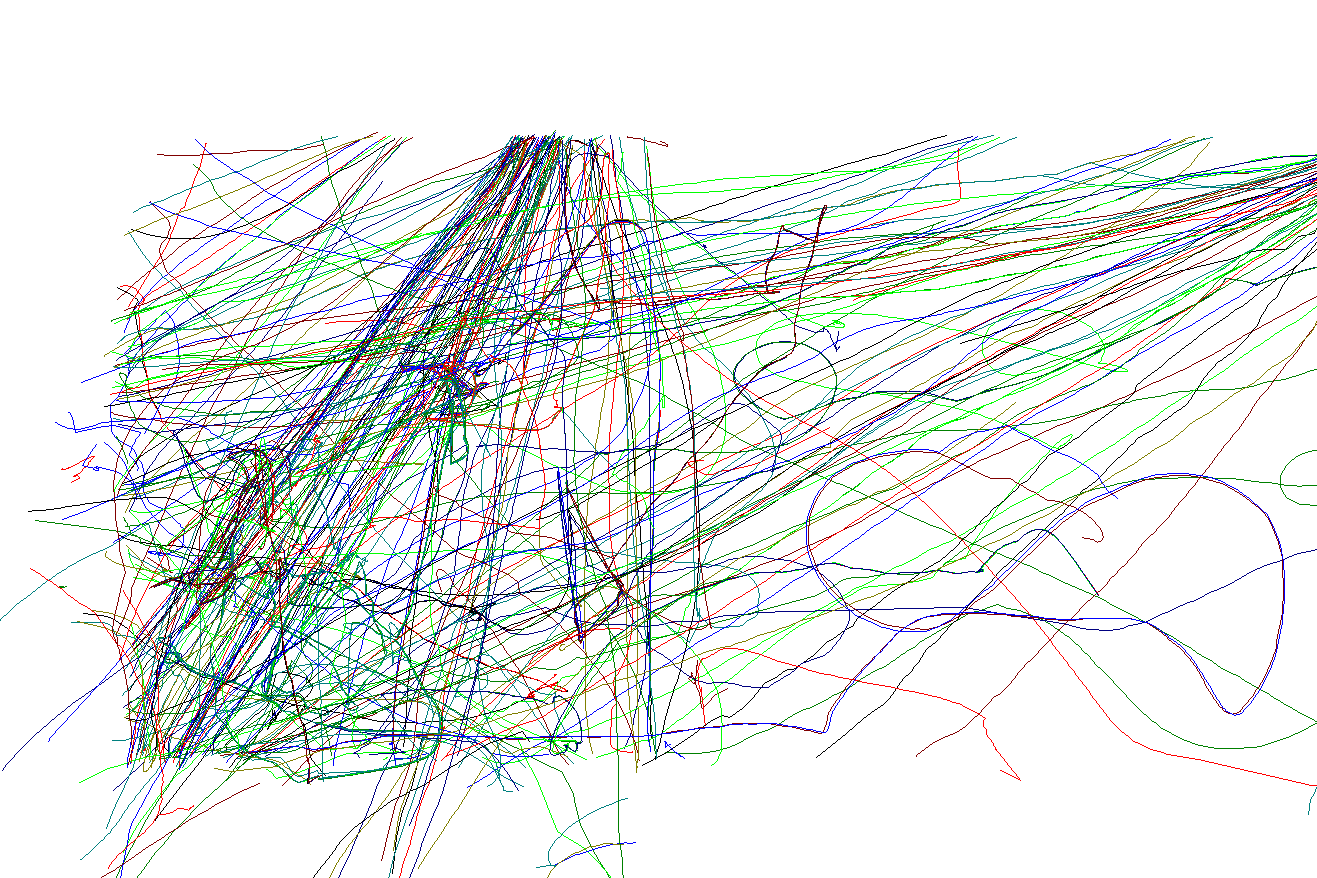}}
\fbox{\includegraphics[width=0.45\textwidth]{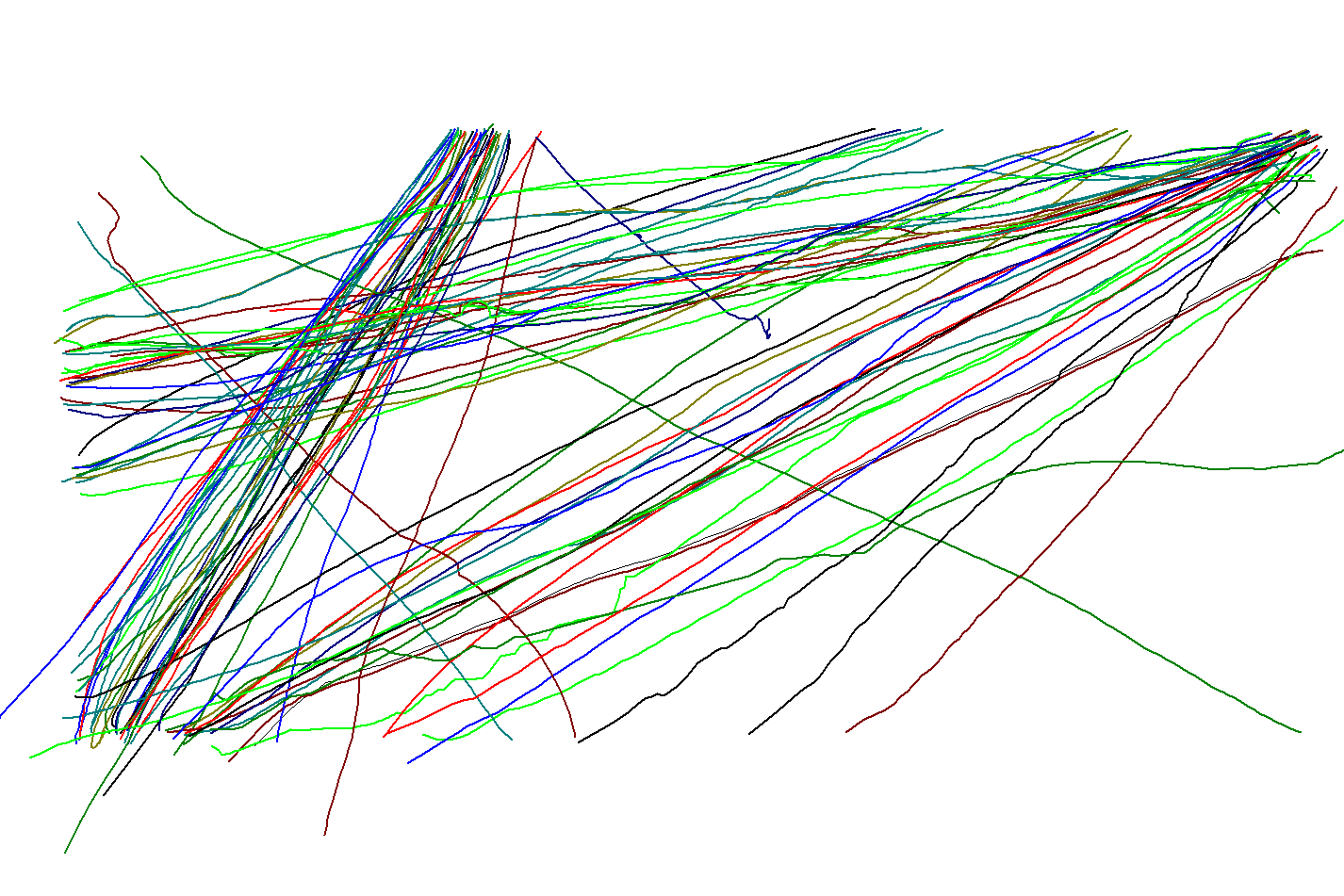}}
\caption{Accumulated tracks from 02-04-2014 (left), extracted normal trajectory (right).}
\label{fig:trajectories1}
\end{figure}

A categorisation can be done with the aid of specific properties such as length, duration or direction of a trajectory. Normal or atypical trajectories can be extrapolated from these features and statistical analyses. A first obvious approach is based on the following criteria:
\textit{Normal trajectories} are the results of direct routes between the entrance door and the doors of different building complexes. An evident criterion is the deviation of the trajectory from the linearity or other smooth curve, which can be described by a polynomial of higher order polynomial (see \autoref{fig:trajectories1} right).

\begin{figure}[t!] 
\centering
\fbox{\includegraphics[width=0.45\textwidth]{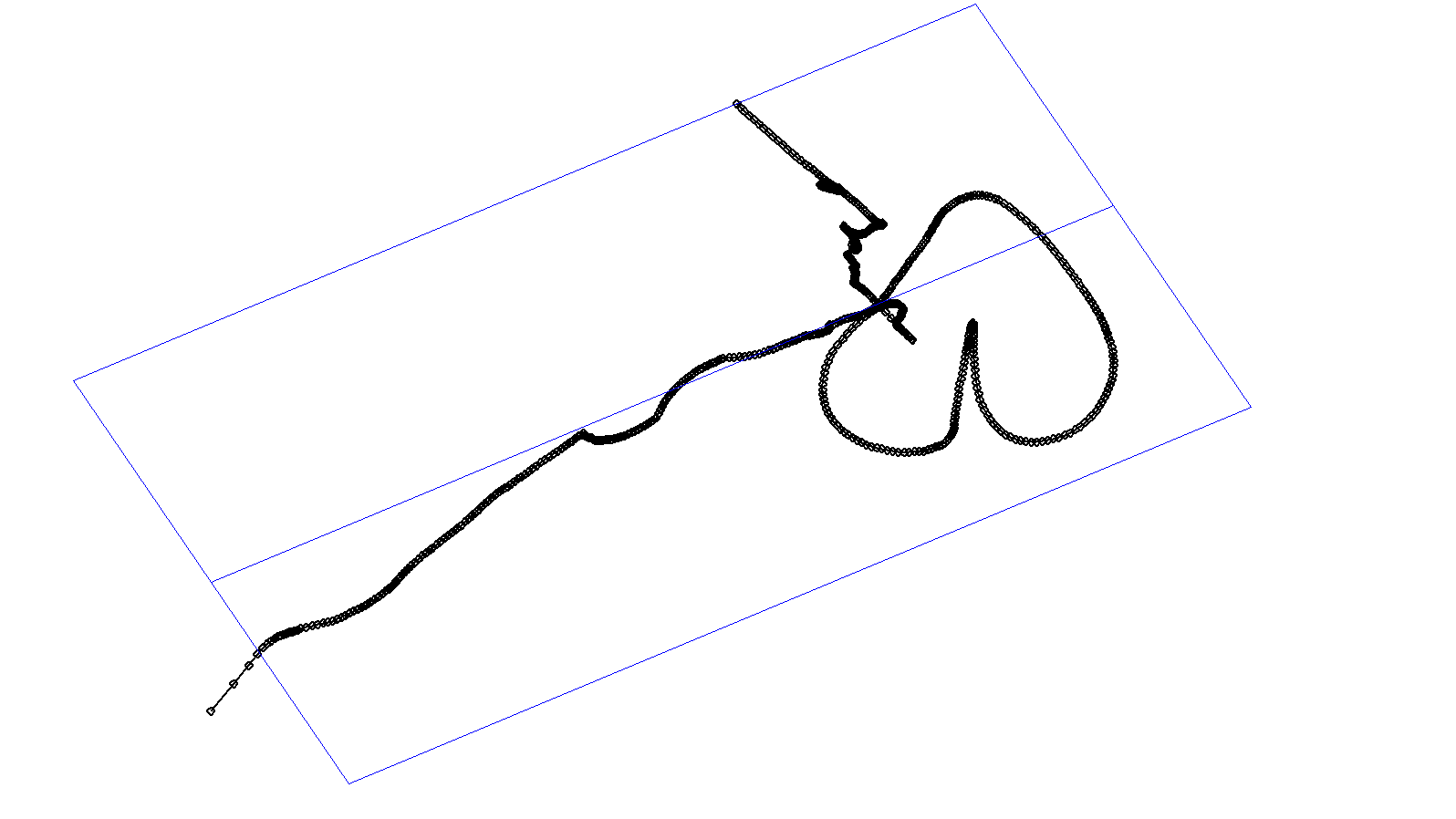}}
\fbox{\includegraphics[width=0.45\textwidth]{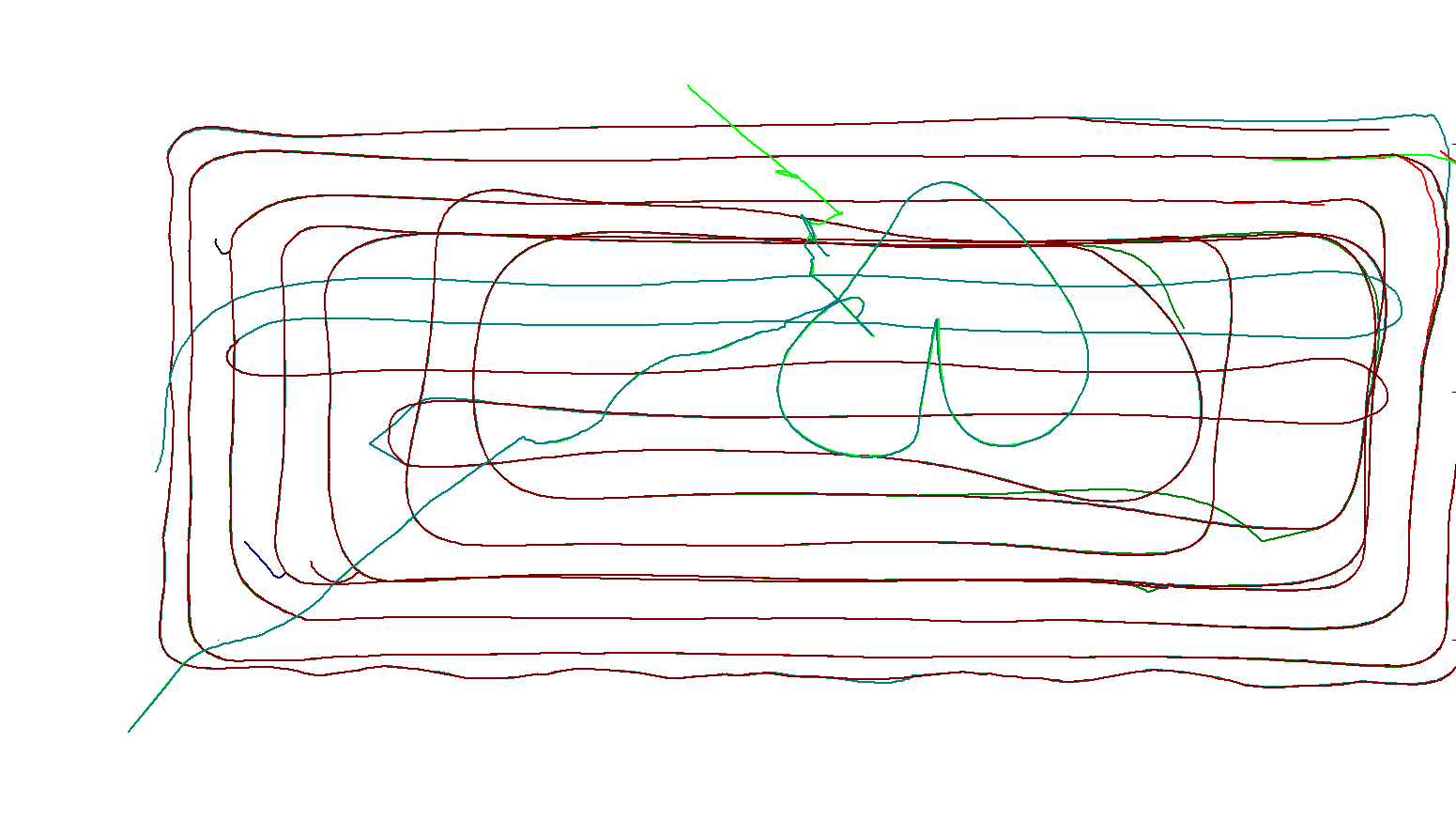}}
\caption{Atypical tracks  (left), collection of atypical trajectory (right).}
\label{fig:trajectories2}
\end{figure}

Based on this, the following criteria are also used for the detection of abnormal objects (see \autoref{fig:trajectories2}): Number of points (nPoints),
 Deviation from the linear fit (dFit),
 Distance travelled (dist),
 Circumference of the figure relative to a horizontal rectangle (cRect) and
 Circumference of the figure relative to a main axis (cMain).

\begin{figure}[t!] 
\centering
\fbox{\includegraphics[width=0.45\textwidth]{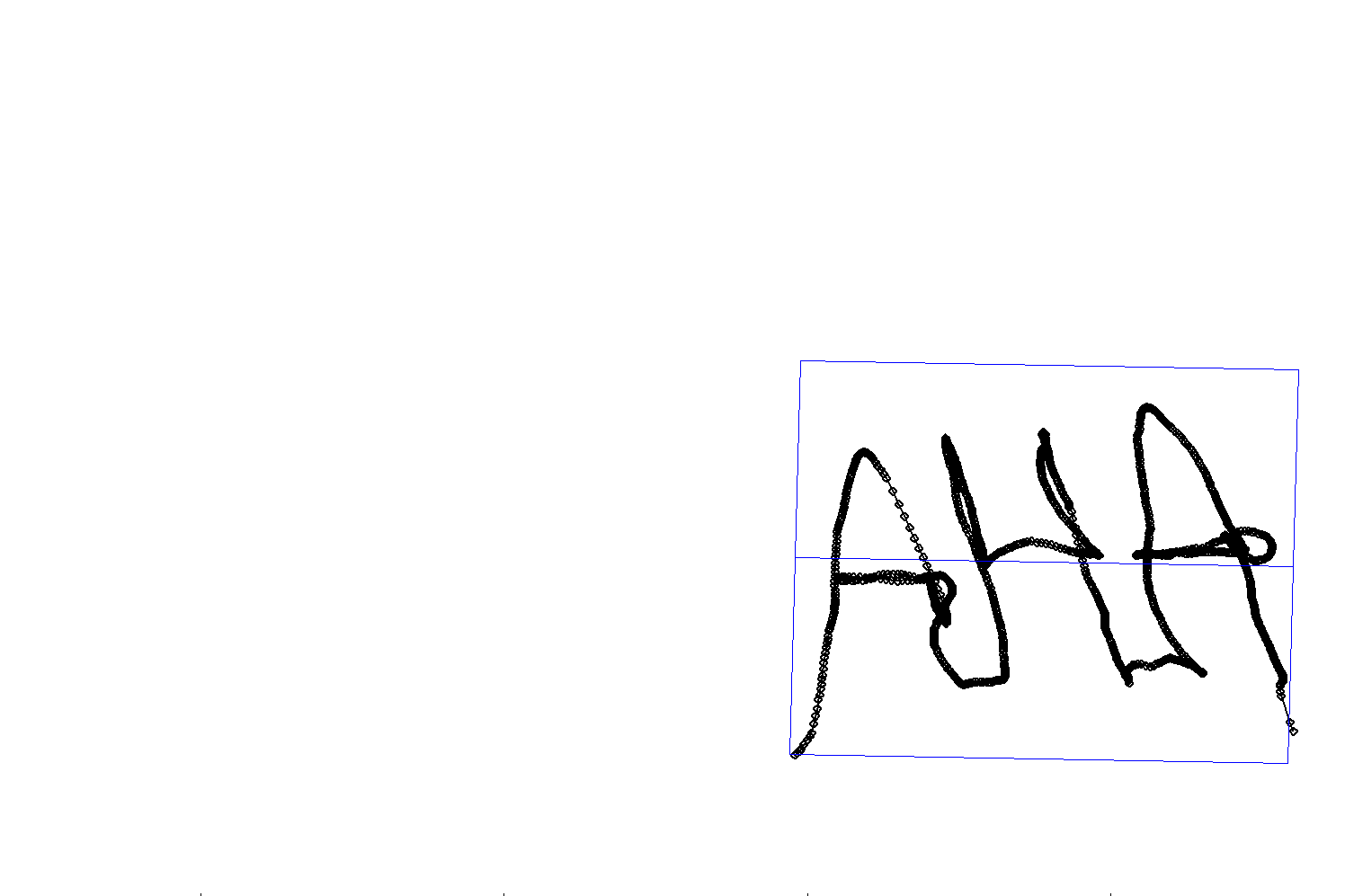}}
\fbox{\includegraphics[width=0.45\textwidth]{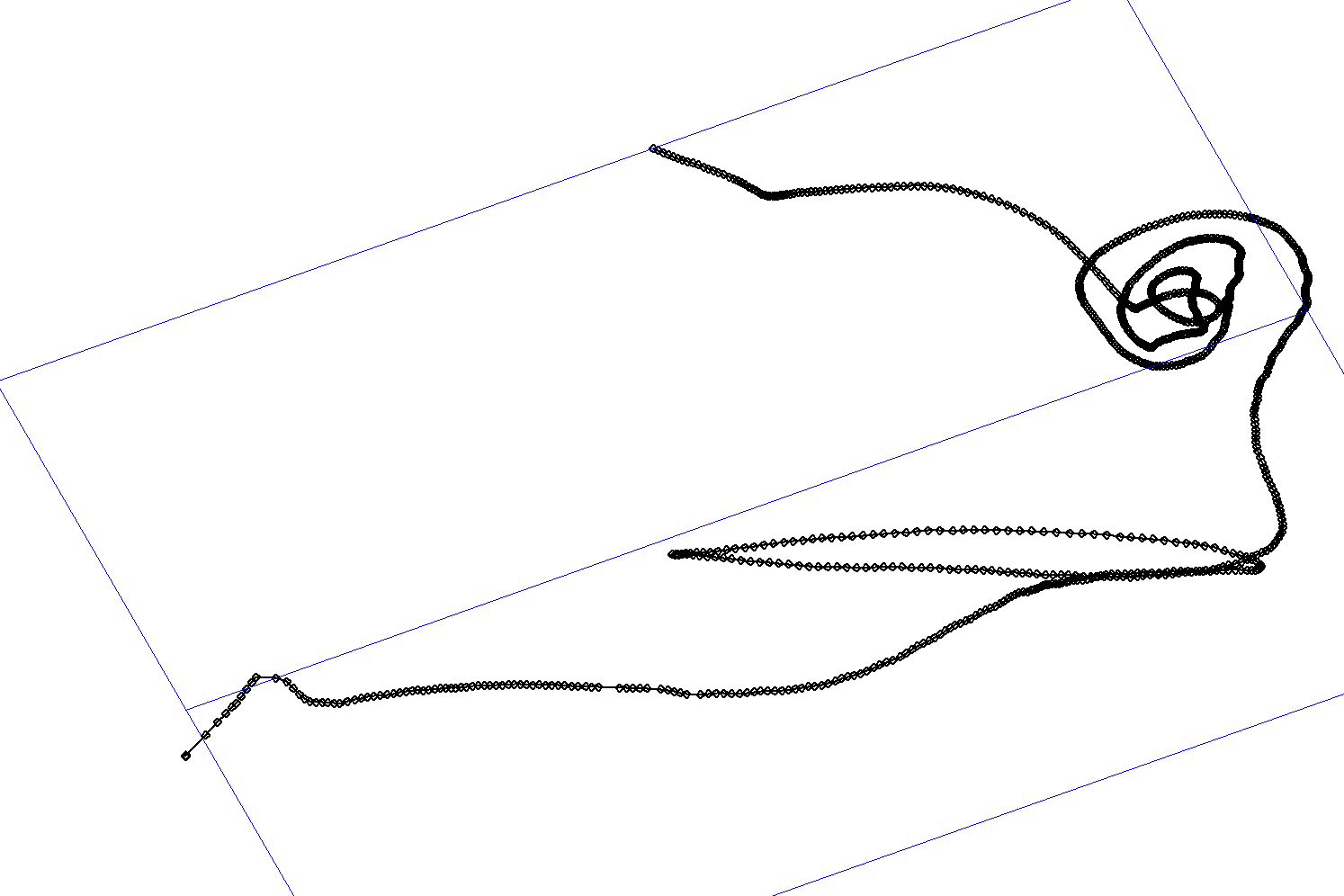}}
\caption{Atypical tracks  (left), collection of atypical trajectory (right).}
\label{fig:trajectories3}
\end{figure}

The \textit{leave a trace} project provides interaction with the system to passersby via a screen in the atrium. Such interactions are expressed by special forms that differ from straight-line movements. For example, a user can write words or draw something artistic. Such a movement pattern is shown in \autoref{fig:trajectories3}.

\subsection{Machine Learning approach for anomaly detection}

Another anomaly detection method was developed on the basis of machine learning. This unsupervised learning method implements a dynamic training set, particularly by using a ring buffer storing trajectory equivalents, in order to provide up-to-date anomaly detection in real time on our data set. This challenge is created by the high variety of events (exhibitions, conferences etc.) that are recorded and have to be handled dynamically. Changing or even just slightly adjusting the training set would require methods like in Johnson and Hogg\cite{RN44} to trigger a new training process, resulting in massive computation power problems at surveillance applications. The same problem applies to the traditional method of manually selecting regions of interests (ROIs) in the camera or top-down view.

Given this dynamic training set, classifying every trajectory as normal or atypical is performed. We define a measure of normality of single trajectories in relation to the training data. Using a threshold the final classification as normal or atypical can be performed. This method allows a real-time classification and even regression for these trajectories that can be used for security applications.

All of the trajectory points were combined in time intervals of 0.5 seconds (trajectory steps). This reduces the amount of data used in the calculation and provides a uniform data quality.

For defining the measure of normality a 4-dimensional so called normality array is defined whose first two dimensions represent the 2D coordinates and whose last two dimensions represent the velocity of the observed persons with its x and y dimension. Each trajectory step can be assigned to a position in this 4-dimensional space.

Every trajectory step in the training data set will increase the value of this array at its corresponding position, increasing the measure of normality of this or a similar step.

The resolution of the 4-dimensional array may be low compared to the original raw data: In our application a (10x10x5x5) size was sufficient. Using this array we are still able to differentiate between 2500 distinct trajectory steps and may calculate the results for steps in between. This requires a transformation from the trajectory space to the normality array space.

In order to apply a trajectory step at its intermediate position in the 4-dimensional array we calculate a Gaussian kernel with its centre at the step's position. The radial basis function kernel is commonly used in machine learning, mainly for support vector machines \cite{RN100}. In this application it can be described as 
\begin{equation} 
K_t(p)=\exp{\left (-\sum_{dim=1}^{4}\frac{(t_{dim}- p_{dim})^2}{\sigma_{dim}^2}  \right )}    ,
\end{equation}
where $t$ is the 4-dimensional trajectory representation and $p$ is the normality array position.
We received good results using (1,1,0.5,0.5) sigma values. Using a kernel instead of a single position respects the differences that exist even between similar movements.

The kernel is added to the 4-dimensional normality array after multiplying it with a weight that is calculated from the current step's velocity. This compensates the stronger influence of very slow movements on the normality array due to a longer time span.

After the normality array has been initialized using the dynamic training data set, the measure of normality of a single step can be calculated as the average of the surrounding array entries weighted with the reciprocal of the distance to these surrounding positions. The measure of normality for trajectory $t$ can therefore be described as

\begin{equation} 
n(t)=\frac{1}{| P_t |}\sum_{p\in P_t}
\frac{N(p)}{\left \| t-p \right \|}
  ,
\end{equation}
where $P_t$ represents the set of the 4-dimensional integer values that surround $t$ (with cardinality $|P_t|$) and $N(p)$ describes the normality array value at $p$.
The normality value of a trajectory can be defined as the average of the normality values of its steps. This defines rare trajectories as atypical while rarity is defined by location and velocity, which can be useful for security and safety applications.

Unloading a trajectory from the normality array in terms of the ring buffer method mentioned above may be performed by subtracting the kernels of its trajectory steps which may be stored in the ring buffer instead of the trajectories respectively.

This method can be executed in real time and requires low computation power compared to other methods used for similar purposes. Another advantage is that the fault tolerance of the method can be matched with the trajectory generation system's specifications (see wobble problem in \autoref{sec:trackingsystemevaluation}) by adjusting the normality array resolution and kernel size and sigma.

\begin{figure}[t!] 
\centering
\includegraphics[width=0.45\textwidth]{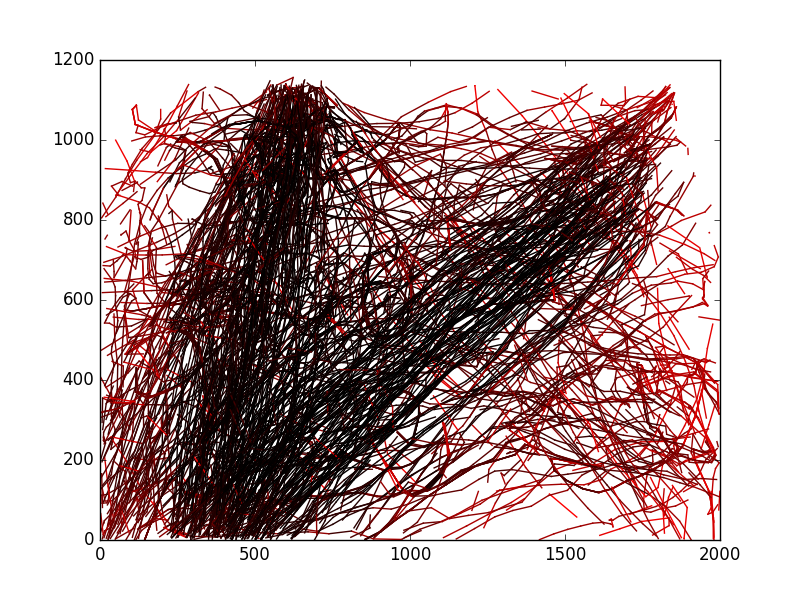}
\includegraphics[width=0.45\textwidth]{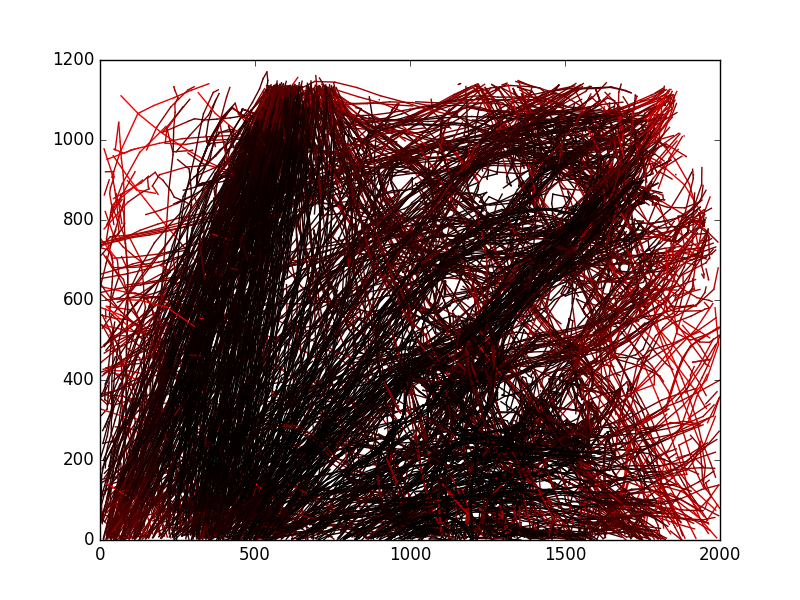}
\caption{All tracks, atypical trajectory are in red. Normal (left) and conference situation (right).}
\label{fig:trajectories5}
\end{figure}

Results can be seen in \autoref{fig:trajectories5}.

In order to perform a final classification the required threshold may be derived by calculating the maximum normality of the most atypical 10\% of the trajectories (or another fixed percentage).

\begin{figure}[t!] 
\begin{center}
\begin{tabular}{cc}
\includegraphics[width=7.5cm]{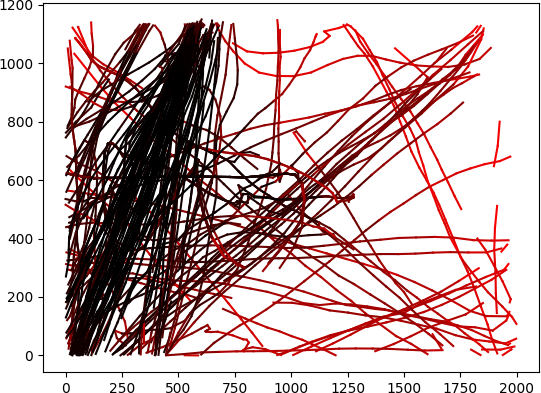}
&\includegraphics[width=7.5cm]{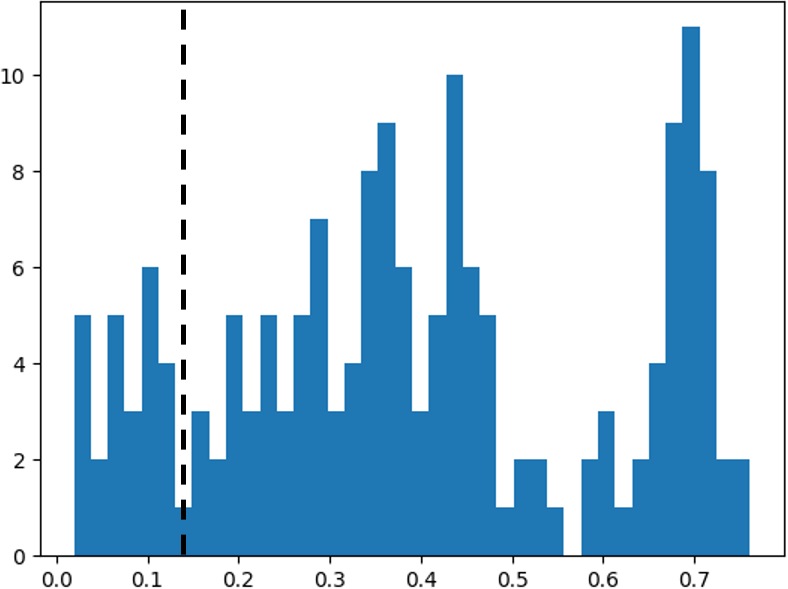}\\(a)&(b)\\
\includegraphics[width=7.5cm]{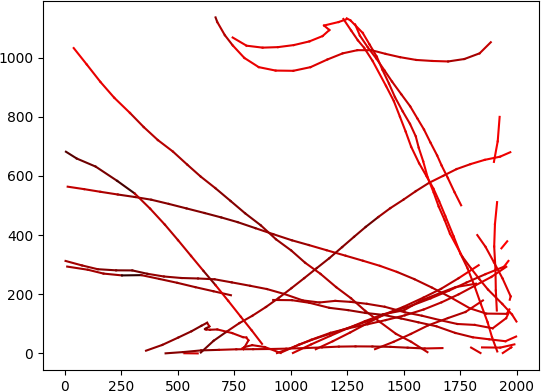}
&\includegraphics[width=7.5cm]{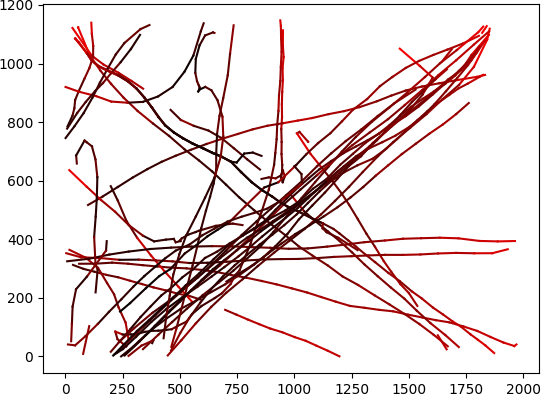}\\(c)&(d)

\end{tabular}
\end{center}
\caption {
(a) Trajectories of everyday scenario with normality depicted by color as in \autoref{fig:trajectories5}. It contains normal and atypical trajectories.
(b) Histogram of trajectories' normality values. Note the gap at 0.14.
(c) Trajectories with a normality below 0.14. They can be identified as atypical.
(d) Trajectories with a normality between 0.14 and 0.3. The result contains no severely atypical trajectories. 
}
\label{fig:hist}
\end{figure}

\autoref{fig:hist} shows a histogram of trajectory normality values of an everyday situation. Because of the nature of normal trajectories they have a normality similar to multiple other trajectories' normality values. This results in higher influence on the histogram which typically leads to a gap between atypical and normal trajectories' values. Maximizing a combination of a histogram's gap descriptor and a function with a peak at the fixed percentage value mentioned above will result in an automatic detection for such classification thresholds.

\section{Conclusion}
We have presented a people tracking system, used for an art in architecture project within an atrium of the Charité in Berlin, Germany. It will let people interact with their own tracks and thus produce individual, artistic results. The visual results are generally different from day to day, which is the artistic concept of this project. 

For this work we had the advantage of having a huge data set at hand. It has been running 9 hours a day, continuously  for the last 3 years and has not produced a system failure yet. The data can be downloaded by everyone, from the Internet-address mentioned in the introduction, which makes this an important contribution to the field of trajectory research. We have also published a script which can open and read the data, again the link for this can be found in the introduction.
This huge data set makes this work very interesting, since it contains many unusual events produced by its users as well as varying scenarios. Also, its shear size forced us to implement robust algorithms which can handle the variety in the appearance of people, objects and the different events held at this atrium. 

The tracking was implemented and optimized for an indoor environment, for difficult lighting situations due to sun reflections and quickly changing illumination conditions. It works in real-time with at least 12 frames per second, with each frame having a size of 2 mega-pixels. The system's hardware consists of one camera placed high above the atrium and a computation unit with up to 8 cores and a fast graphics card. This poses a difficult problem, in future projects it might be easier to save money on other hardware and introduce more sensors, like a second camera. This would allow for more precise tracking, even of persons who wear a white coat which is pretty much the color of the background.

In the future the system can be improved by more robust object detection and better group handling (clustering). Since many applications will use only one sensor or camera, the focus will likely stay within the mono-camera applications.

In this paper we proved, that machine learning approaches should be applied in order to process this data. The proposed approach is able to handle a variety of different scenarios because of a dynamic training set, which makes this approach novel and suitable in the area of trajectory analysis. It may be extended further using different machine learning methods with the goal of developing a real-time adapting classification which will be a topic of further research.

\acknowledgments
We would like to thank Prof. Tyyne Claudia Pollmann and the Charité for making the track data available to us and the public. Finally, we would also like to thank Alberto Massa and Marc Uhlig for their contribution to the implementation of the tracking system and data evaluation process.

\end{spacing}

\end{document}